\documentclass[letterpaper, 10 pt, conference]{ieeeconf}  

\IEEEoverridecommandlockouts                              

\usepackage{stfloats}
\usepackage{bbm}

\usepackage{graphicx} 
\usepackage{amsmath} 
\usepackage{amssymb}  
\let\labelindent\relax
\usepackage{enumitem}
\usepackage{lipsum}
\usepackage{graphicx}
\usepackage[T1]{fontenc}
\usepackage{aecompl}
\usepackage{amsfonts}
\usepackage{bbm}
\usepackage{subcaption}
\usepackage{ifthen}
\usepackage{cite}
\usepackage[usenames, dvipsnames]{color}
\usepackage{url}
\usepackage{hyperref}
\usepackage[hyphenbreaks]{breakurl}
\usepackage{mathtools}	
\usepackage[normalem]{ulem}
\usepackage{algorithm2e}
\RestyleAlgo{ruled}
\usepackage{algpseudocode}
\usepackage{tabularx}

\usepackage{amsthm}
\usepackage{balance}

\newtheorem{theorem}{Theorem}
\newtheorem{lemma}{Lemma}
\newtheorem{definition}{Definition}
\newtheorem{corollary}{Corollary}

\newtheorem{remark}{Remark}
\newtheorem{assumption}{Assumption}

\DeclareMathOperator{\vol}{vol}
\newcommand{\norm}[1]{\left\lVert#1\right\rVert}

\DeclareMathOperator*{\argmax}{arg\,max}
\usepackage{bbm}

\newcommand{\A}{\mathcal{A}}
\newcommand{\X}{\mathcal{X}}
\newcommand{\R}{\mathcal{R}}
\DeclareMathOperator{\opint}{int}

\makeatletter
\newcommand*{\centerfloat}{%
  \parindent \z@
  \leftskip \z@ \@plus 1fil \@minus \textwidth
  \rightskip\leftskip
  \parfillskip \z@skip}
\makeatother

\usepackage{ifthen}
\newboolean{showcomments}
\setboolean{showcomments}{true}
\usepackage[usenames,dvipsnames]{color}
\usepackage{todonotes}

\definecolor{bleudefrance}{rgb}{0.19, 0.55, 0.91}
\definecolor{ao(english)}{rgb}{0.0, 0.5, 0.0}

\newcommand{\addcite}[0]{\ifthenelse{\boolean{showcomments}}
{\textcolor{purple}{(add cite(s)) }}{}}%

\newcommand{\enrique}[1]{  \ifthenelse{\boolean{showcomments}}
{\todo[inline,color=bleudefrance]{Enrique: #1}}{}}
\newcommand{\emmargin}[1]{\ifthenelse{\boolean{showcomments}}{\marginpar{\color{bleudefrance}\tiny EM: #1}}{}}

\newboolean{showedits}
\setboolean{showedits}{false}
\usepackage[markup=underlined]{changes}
\definechangesauthor[color=bleudefrance]{EM}
\newcommand{\aem}[1]{
\ifthenelse{\boolean{showedits}}
{\added[id=EM]{#1}}
{\!#1\hspace{-4.75pt}}
}
\newcommand{\repem}[2]{
\ifthenelse{\boolean{showedits}}
{\replaced[id=EM]{#1}{#2}}
{\!#1\hspace{-4.75pt}}
}
\newcommand{\dem}[1]{
\ifthenelse{\boolean{showedits}}
{\deleted[id=EM]{#1}}
{}
}

\makeatletter
\if@todonotes@disabled

\else

\fi
\makeatother

\newboolean{with-appendix}
\setboolean{with-appendix}{false}
\mathtoolsset{showonlyrefs=true}

\title{Model-free Learning of Regions of Attraction via Recurrent Sets}

\author{Yue Shen, 
Maxim Bichuch, and Enrique Mallada 
\thanks{
Y. Shen and E. Mallada are with the Department of Electrical and Computer Engineering and M. Bichuch is with the Department of Applied Mathematics and Statistics, Johns Hopkins University, 3400 N Charles St., Baltimore, MD 21218, USA \texttt{\{yshen50,mbichuch,mallada\}@jhu.edu}.}}

\begin{document}

\allowdisplaybreaks

\maketitle
\begin{abstract}
We consider the problem of learning an inner approximation of the region of attraction (ROA) of an asymptotically stable equilibrium point without an explicit model of the dynamics. Rather than leveraging approximate models with bounded uncertainty to find a (robust) invariant set contained in the ROA, we propose to learn sets that satisfy a more relaxed notion of containment known as recurrence. We define a set to be $\tau$-recurrent (resp. $k$-recurrent) if every trajectory that starts within the set, returns to it after at most $\tau$ seconds (resp. $k$ steps). We show that under mild assumptions a $\tau$-recurrent set containing a stable equilibrium must be a subset of its ROA. We then leverage this property to develop algorithms that compute inner approximations of the ROA using  counter-examples of recurrence that are obtained by sampling finite-length trajectories. Our algorithms process samples sequentially, which allows them to continue being executed even after an initial offline training stage. We further provide an upper bound on the number of counter-examples used by the algorithm, and almost sure convergence guarantees. 
\end{abstract}

\medskip


\section{Introduction}\label{sec:intro}
The problem of estimating the region of attraction (ROA) of an asymptotically stable equilibrium point has a long-standing history in nonlinear control and dynamical systems theory \cite{Khalil2002}.
From a theoretical standpoint, there has been a thorough study of conditions that guarantee several topological properties of such set, e.g., being connected, open, dense, smooth \cite{willard2004general}.
From a practical standpoint, having a representation of such region allows to test the limits of controller designs, which are usually based on (possibly linear) approximations of nonlinear systems~\cite{li2020online}, and provides a mechanism for safety verification of certain operating conditions~\cite{robey2020learning} \cite{Sallab2017deep}.
Unfortunately, it is known that finding an analytic form of the region of attraction is difficult and in general impossible \cite[p. 122]{Khalil2002}. 
As a result, most efforts in characterizing the ROA focus on finding inner approximations by means of invariant sets.

\subsubsection{Related Work} 
Several methodologies for computing inner approximations of the ROA have been proposed in the literature. In a broad sense, they can be classified into three groups, depending on whether accurate, inaccurate, or no information about the dynamic model is present. Notably, at their core, almost all of the methods rely on finding an invariant set of the system. We briefly review such methods next.

\noindent
\textit{Exact Models:}
When an exact description of the dynamics is available, it is possible to use this information via two complementary methodologies. Lyapunov methods utilize the fact that Lyapunov functions are certificates of asymptotic stability and build inner approximations using its sublevel sets.  Methods for finding such Lyapunov functions are surveyed in, e.g., \cite{Peter2015}. In particular, \cite{VANNELLI198569} and \cite{HASSAN1981} construct Lyapunov functions that are solutions of Zubov's equation, and  \cite{Pedro1999}  searches for piece-wise linear Lyapunov functions that are found via linear programming. Similarly,  piece-wise quadratic parameterizations of Lyapunov functions using LMI-based  methods are considered in \cite{Goebel2006}. Finally, recent work \cite{chen2020learning} leverages the universal approximation property of neural networks to estimate the ROA of general nonlinear dynamical systems. Alternatively, non-Lyapunov methods focus directly on the properties of the ROA. For example, trajectory reversing methods \cite{Gensio1985} \cite{chiang1988} derive the boundary of ROA directly from the stable manifold of the equilibria on the boundary, and the reachable set method \cite{Baier2009} generates a grid of sample points and classifies each of them by solving an optimal control problem. 

\noindent
\textit{Inexact Models:}
In the presence of uncertainty, robust ROA approximation methods \cite{xue2020robust,Ambrosino2012,chen2021learning,topcu2009robust} generalize Lyapunov approaches by finding a common Lyapunov function across the entire uncertainty set.
Alternatively, learning-based methods utilize experimental data to estimate the region of attraction. When a Lyapunov function is provided, experimental data expand the Lyapunov function level set through, e.g., Gaussian processes \cite{berkenkamp2016safe}, or a simple sampling approach \cite{Najafi2016}. To address the problem of simultaneously learning the Lyapunov function and the level set, \cite{richards2018lyapunov} parameterizes the Lyapunov function as a neural network and iteratively trains it by sampling points that are outside of the current Lyapunov level set but come back in within $T$ steps.

\noindent
\textit{Model-free:}
Notably, learning methods play a crucial role in model-free settings. In particular, similar to the Lyapunov methods, \cite{Colbert2018EstimatingTR} uses trajectory data to fit values of a Lyapunov function by leveraging converse Lyapunov results. Perhaps most relevant to our paper is \cite{wang2020modelfree}, which establishes a non-Lyapunov approach that determines the boundary of ROA directly from a support vector machine, trained from experimental data that is sampled via hybrid active learning techniques. 


\subsubsection{Contributions}
In this paper, we provide a novel approach for learning inner approximations of the region of attraction of an asymptotically stable equilibrium point from sampled finite-length trajectories. 
\textcolor{black}{
We refer to such a method as ``model-free'' since it does not require an explicit description of the system but only requires a process that generates the sample trajectories.
}

Rather than focusing on learning invariant sets that require trajectories to always lie within the set, we propose to learn sets that satisfy a more flexible notion of invariance. The contributions of this work are manifold:
\begin{itemize}
    \item We propose the notion of recurrence as an alternative property that can be used to guarantee a set to be contained in the region of attraction. 
    \item We show that under mild conditions, a compact set containing an asymptotically stable equilibrium point is a subset of the region of attraction if and only if it is recurrent.
    \item We leverage this property to develop several algorithms that can learn inner approximations of the region of attraction using counter-examples of recurrence that are based on finite-length  trajectory samples.
    \item We further provide guarantees on the worst-case number of counter-examples required to compute a recurrent set.
\end{itemize}

\subsubsection{Organization}
The rest of the paper is organized as follows. In Section \ref{sec:formulation}, we formulate the problem we aim to solve, as well as revisit some classical results that will be leveraged in this work. The notion of recurrence to be used in this work is introduced in Section \ref{sec:recurrent}, together with our first core set of results that show the relationship between recurrence and containment within the region of attraction. The proposed algorithms and the corresponding guarantees are given in Section \ref{sec:Solution Approach}. Numerical examples are provided in Section \ref{sec:experiments} and we conclude in Section \ref{sec:conclusions}.

\ifthenelse{\boolean{with-appendix}}{
\textcolor{black}{
\subsubsection{Proofs}  We provide detailed proofs for all our results in the online version of this work \cite{paper-w-proofs}.
}
}{}
\section{Problem Formulation}\label{sec:formulation}
We consider a continuous time dynamical system 
\begin{equation}\label{eq:system}
 \dot{x}(t) = f(x(t))\,,  
\end{equation}
where $x(t)\in\mathbb{R}^d$ is the state at time $t$, and the map $f: \mathbb{R}^d\rightarrow \mathbb{R}^d$ is continuously differentiable and (globally) Lipschitz. Given initial condition $x(0) = x_0$, we use $\phi(t,x_0)$ to denote the solution of \eqref{eq:system}. Using this notation, the positive orbit of $x_0$ is given by $\mathcal{O}_{+}(x_0) = \{ y\in \mathbb{R}^d:\, y=\phi(t,x_0),\, t\in\mathbb{R}^+\}$.

\begin{definition}[$\omega$-limit Set]\label{defn:omega-limit-set}
Given an initial condition $x_0$, its $\omega$-limit set $\Omega(x_0)$ is the set of points $y\in\mathbb{R}^d$ for which there exists a sequence $t_n$ indexed by $n\in\mathbb{N}$ satisfying $\lim_{n\rightarrow\infty} t_n =\infty$ and $\lim_{n\rightarrow\infty} \phi(t_n,x_0) =y$.  We will further use $\Omega(f)$ to denote the $\omega$-limit set of \eqref{eq:system}, which is the union of $\omega$-limit sets of all $x\in\mathbb{R}^d$.
\end{definition}
 Note that by definition, if $x^*$ is an equilibrium  of \eqref{eq:system}, then it follows that $x^*\in\Omega(f)$. 

\subsection{Region of Attraction}
We would like then to learn the set of initial conditions that converge to $x^*$.
\begin{definition}[Region of Attraction]\label{defn:ROA}
\textcolor{black}{
Given an invariant set $S\subseteq\Omega(f)$, the region of attraction (ROA) of $S$ under \eqref{eq:system} is defined as
\begin{align}\label{eq:ROA}
\mathcal{A}(S) := \left\{x_0 \in\mathbb{R}^d|\liminf_{t\rightarrow\infty} d(\phi(t,x_0), S )=0\right\},
\end{align}
where $d(y,S)$ is the distance from the solution  $y$ to the set $S$, i.e., $d(y,S):= \min_{x\in S}\norm{x-y}_2$. When the set $S$ is a singleton that contains exactly one point (say $x$), we abbreviate $\mathcal{A}(S) = \mathcal{A}(\{x\})$ as $\mathcal{A}(x)$.
}
\end{definition}

Note that without further assumptions, the set \eqref{eq:ROA} may be a singleton, have zero measure, or be disconnected, making the problem of characterizing \eqref{eq:ROA} from samples almost impossible. We thus make the following assumption.

\begin{assumption}\label{as:roa}
The system \eqref{eq:system} has an asymptotically stable equilibrium at $x^*\in\mathbb{R}^d$.
\end{assumption}

\begin{remark}
It follows from Assumption~\ref{as:roa} that the ROA $\mathcal{A}(x^*)$  is an open contractible set~\cite{sontag2013mathematical}, i.e., the identity map of $\mathcal{A}(x^*)$ to itself
is null-homotopic~\cite{Munkres2000}.
\end{remark}

Having set up the necessary assumption for an ROA to be learnable, we now move on to a certain property that helps us to characterize subsets of the region of attraction. 

By definition, $\mathcal{A}(S)$ satisfies the invariant property that every trajectory that starts in the set $\mathcal{A}(S)$ remains in the set for all future times, i.e., $\mathcal{A}(S)$ is a positively invariant set~\cite{Khalil2002}.

\begin{definition}[Positively Invariant Set]\label{defn:invariant set}
A set $\mathcal{I}\subseteq \mathbb{R}^d$ is positively invariant w.r.t. \eqref{eq:system} if and only if:
\begin{equation}\label{eq:invariant-criterion}
x_0\in \mathcal{I}\implies \phi(t,x_0)\in \mathcal{I},\quad \forall\, t\in \mathbb{R}^+.
\end{equation}
\end{definition}
The notion of positive invariance is fundamental for control. It is used to trap trajectories in compact sets and allows the development of the Lyapunov theory. By trapping trajectories on sub-level sets of a function, one can guarantee boundedness of trajectories, stability, and even asymptotic stability via a gradual reduction of the value of the Lyapunov function. 
A natural approach  is therefore to search for Lyapunov functions~\cite{Khalil2002} that render its sublevel sets as invariant inner-approximations of $\mathcal{A}(x^*)$. Such methods are particularly justified after the fundamental result by Vladimir Zubov~\cite{driver1965methods} that guarantees the existence of such a function:
\begin{theorem}[Zubov's Existence Criterion]\label{thm:zubov}
A set $\A$ containing $x^*$ in its interior is the region of attraction of $x^*$ under \eqref{eq:system} if and only if there exist continuous functions $V$, $h$ such that the following hold:
\begin{itemize}
    \item $V(x^*)=h(x^*)=0$, $0<V(x)<1$ for $x\in\A\backslash\{x^*\}$, $h(x)>0$ for $x\in\mathbb{R}^d\backslash\{x^*\}$.
    \item For every $\gamma_2>0$, there exists $\gamma_1>0$, $\alpha_1>0$ such that
    $V(x)>\gamma_1$, $h(x)>\alpha_1$,  whenever $\|x\|\geq \gamma_2$. 
    \item $V(x_k)\rightarrow1$ for all sequences $\{x_k\}$ such that $x_k\rightarrow\partial \A$ or $\|x_k\|\rightarrow\infty$.
    \item $V$ and $h$ satisfy
    \begin{equation}\label{eq:dotVF}
        (\mathcal{L}_fV)(x) = -h(x)(1-V(x))\sqrt{1+\|f(x)\|^2},
    \end{equation}
    where $(\mathcal{L}_fV)(x)$ is the Lie derivative of $V$ under the flow induced by $f$.
\end{itemize}
    
\noindent
Particularly, when $f(x)$ is continuously differentiable, $h(x)$ can always be selected such that $V$ is differentiable, i.e.,  $(\mathcal{L}_fV)(x)=\nabla V(x)^Tf(x)$.
\end{theorem}

\begin{corollary}\label{cor:level set connetness}
Under Assumption \ref{as:roa}, there exists a Lyapunov function $V$ with domain on $\A(x^*)$ such that for any $c\in(0,1)$ the sublevel set $V_{\leq c}:=\{x:V(x)\leq c\}$ is a contractible invariant subset of $\A(x^*)$. 
\end{corollary}
\ifthenelse{\boolean{with-appendix}}{
}{
\begin{proof}
Let $V$ be the Zubov's function whose existence is guaranteed by Theorem \ref{thm:zubov}. Thus by the definition of $V$, for $c\in(0,1)$, $V_{\le c}\subseteq\mathcal{A}(x^*)$. Further from \eqref{eq:dotVF}, it follows that $(\mathcal{L}_fV)(x)\le 0$, for $x\in V_{\le c}\subset \A(x^*)$. Thus, $V_{\le c}$ is positively invariant.

To prove the $V_{\le c}$ is contractible, we need to provide a continuous mapping $H:[0,1]\times V_{\le c} \rightarrow V_{\le c}$ such that $H(0,x)=x$ and $H(1,x)=x^*$ for all $x\in V_{\le c}$. Similar to~\cite{sontag2013mathematical}, we define $H(s,x):= \phi(\frac{s}{1-s},x)$ for $s<1$, and $H(1,x)\equiv x^*$. Note that $H$ is continuous in $s$ and $x$ for $s<1$ , as in ~\cite{Khalil2002}. 
We are thus left to prove continuity at each $(1,x)$. 
To do so, we take any such $x$ and pick any open neighborhood $\mathcal{V}$ of $H(1,x)=x^*$. By Assumption~\ref{as:roa} as well as the definition of asymptotic stability, it follows that there exists another open neighborhood $\mathcal{W}\subseteq\A(x^*)$ of $x^*$ for which all trajectories starting in $\mathcal{W}$ remain in $\mathcal{V}$, i.e., $\phi(t,x_0)\in\mathcal{V}$ for all $x_0\in\mathcal{W}$ and $t>0$. Given $V_{\le c}\subseteq\mathcal{A}(x^*)$, any point $x\in V_{\le c}$ satisfies $\phi(T,x)\in\mathcal{W}$ for some $T>0$. This, together with the continuity of $\phi(T,\cdot)$, implies that there is a neighborhood $\mathcal{V}'\subseteq V_{\le c}$ of $x$ such that $\phi(T,y)\in\mathcal{W}$ for all $y\in \mathcal{V}'$, which let us conclude:
\begin{align}
    H(s,y)\in \mathcal{V} \quad \text{whenever}\,\, y\in\mathcal{V}'\,\,\text{and}\,\, s>1-\frac{1}{T+1}
\end{align}
and continuity follows since $\mathcal{V}$ could be made arbitrarily small.
\end{proof}
}

The Zubov's function $V$ of Theorem \ref{thm:zubov} provides a parametric family $\{V_{\leq c}:c\in(0,1)\}$ of positively invariant sets inside $\A(x^*)$. Further, while Zubov's result provides a constructive method for $V(x)$, by means of solving a partial differential equation, such a method becomes impractical in the absence of a descriptive model for \eqref{eq:system}. 
Thus, in the absence of an exact model of the dynamics, it is natural to try to find a set inside $\A(x^*)$ that is positively invariant in a robust sense, in the presence of bounded uncertainty~\cite{topcu2009robust}, or that is positively invariant with high probability~\cite{berkenkamp2016safe}. 

However, one of the caveats of positively invariant sets is that they need to be specified very carefully, in the sense that even a good approximation of a positively invariant set is not necessarily positively invariant. Particularly, subsets of positively invariant sets need not be positively invariant.
This indirectly imposes strict constraints on the complexity of the set that one needs to learn via \eqref{eq:invariant-criterion}.
This motivates the alternative proposed in the next section.

\section{Recurrent Sets}\label{sec:recurrent}
We now introduce the relaxed notion of invariance to be used in this paper, which we refer to here as recurrence.
We will then illustrate how recurrent sets constitute a more flexible and more general class of objects of study.
\textcolor{black}{
\begin{definition}[Recurrent Set]\label{defn:recurrent}
A set $\mathcal{R} \subseteq \mathbb{R}^d$ is recurrent w.r.t. \eqref{eq:system}, if for any point $x_0\in \R$ and any time $t\ge 0$, there exists a time $t'> t$, such that $\phi(t',x_0)\in \R$.
\end{definition}
}

\textcolor{black}{Note that a recurrent set, while not invariant, guarantees that solutions starting in this set will visit it back infinitely often. In particular, by Definition \ref{defn:invariant set}, a positively invariant set $\mathcal{I}$ is recurrent.
Thus, } Definition \ref{defn:recurrent} generalizes the notion of positive invariance by allowing the solution $\phi(t,x_0)$ to step outside the set $\mathcal{R}$ for some finite time. Moreover, in what follows, we do not make assumptions on the connectivity of $\mathcal{R}$, and thus $\mathcal{R}$ could be disconnected to better approximate the ROA.
One concern may be however that by allowing $\phi(t,x_0)$ to leave the set $\mathcal{R}$, this will lead to trajectories that diverge, thus leading to unstable behavior. The following \textcolor{black}{result} shows that under mild assumptions, this should not be a source of concern.

{\color{black}
\begin{lemma}\label{lem:bounded traj}
\textcolor{black}{Let $\R\subset\mathbb{R}^d$ be a compact recurrent set satisfying $\partial\R\cap\Omega(f)=\emptyset$.
Then for any $x_0\in\R$, there exists some time  $T>0$, such that the solution $\phi(t,x_0)\in\mathcal{R}$ for all $t\ge T$.}
\end{lemma}

\begin{proof}
    We will prove this statement by contradiction. Assume the result does not hold, i.e., there exists $x_0\in\mathcal{R}$ s.t. for any $t > 0$ there exists a $t'\ge t$ such that $\phi(t',x_0)\not\in\mathcal{R}$. This, together with the definition of the recurrent set (Definition~\ref{defn:recurrent}) and the continuity of the solution, implies there exists a $t''\ge t$ such that $\phi(t'',x_0)\in\partial\mathcal{R}$ for any $t>0$. Therefore, we can construct an infinite sequence $\{x_{n}\}_{n=0}^\infty$ that lies within $\partial\R$, i.e., $\{x_{n}\}_{n=0}^\infty\subset\partial\R$.

    Precisely, let $t_0\ge0$ be a time such that $\phi(t_0,x_0)\in\partial\R$. Then, given $x_{n}:=\phi(t_n,x_0)\in\partial\R$ and some fixed time interval $\tau>0$, we defined $t_{n+1}$ as the first time since $t_n+\tau$ that the solution $x_{n+1}:=\phi(t_{n+1},x_0)$ lies within $\partial\R$, i.e.,  $\phi(t_{n+1},x_0)\in\partial\R$ and $\phi(t,x_0)\not\in\partial\R$ for all $t\in[t_n+\tau,t_{n+1})$.

    Then, since $\partial\R$ is compact, by Bolzano-Weierstrass theorem, $\{x_{n}\}_{n=0}^\infty$ must have a sub-sequence  $\{x_{n_i}\}_{i=1}^\infty$ that converges to an accumulation point $\bar x\in \partial\R$. It follows then from the definition of $\omega$-limit sets (Definition \ref{defn:omega-limit-set}) that $\bar x= \lim_{i\rightarrow \infty}x_{n_i}\in \Omega(f)\cap\partial\R$, which contradicts with the assumption that $\partial\R\cap\Omega(f)=\emptyset$.
\end{proof}

After characterizing regularity conditions for trajectories starting from a recurrent set $\mathcal{R}$, we are ready to show how recurrent sets can be used to characterize subsets of an ROA.

\begin{theorem}\label{thm:recurrence}
Let $\R\subset\mathbb{R}^d$ be a compact set satisfying $\partial\R\cap\Omega(f)=\emptyset$.
Then $\R$ is recurrent if and only if $\,\Omega(f)\cap\R\not=\emptyset$ and $\R\subset\A(\Omega(f)\cap\R)$.
\end{theorem}

\begin{proof}


($\implies$):~
If $\R$ is a compact recurrent set satisfying $\partial\R\cap\Omega(f)=\emptyset$, Lemma~\ref{lem:bounded traj} implies that for any point $x_0\in \R$, there exists a time $T>0$ such that $\phi(t,x_0)\in\mathcal{R} ,\,\forall t\ge T$, i.e., the solution is bounded in the compact set $\R$ for all $t\ge T$. It then follows from \cite[p. 127]{Khalil2002} that the limit set $\Omega(x_0)\not=\emptyset$ and $\lim_{t\rightarrow\infty} d(\phi(t,x_0),\Omega(x_0))=0$. Therefore, we conclude $\Omega(f)\cap\R\supseteq\Omega(x_0)\not=\emptyset$ and $x_0\in\A(\Omega(f)\cap\R)$. Finally, since $x_0$ was chosen arbitrarily within $\R$, it follows that $\R\subset\A(\Omega(f)\cap\R)$.

\noindent
($\Longleftarrow$):~
By assumption $\Omega(f)\cap\R\subset \opint\R$. Therefore, we can always construct an open $\zeta$-neighborhood ${\Omega}_\zeta^\R:=\{x\in\mathbb{R}^d\vert d(x,\Omega(f)\cap\R)<\zeta\}$ of $\Omega(f)\cap\R$ for some $\zeta>0$ small enough such that ${\Omega}^\R_\zeta\subset\opint\R$.

Then for any point $x_0\in\R$, by the assumption that $\R \subset \A(\Omega(f)\cap\R)$, the solution $\phi(t,x_0)$ converges to $\Omega(f)\cap\R$, i.e., $\liminf_{t\rightarrow\infty}d(\phi(t,x_0),\Omega(f)\cap\R)=0$. It follows then that for any $\zeta>0$ and time $t>0$, there always exists some time $t'\ge t$ such that $d(\phi(t',x_0),\Omega(f)\cap\R)<\zeta$, and thus $\phi(t',x_0)\in{\Omega}^\R_\zeta\subset\R$. Therefore, $\R$ is recurrent.
\end{proof}
}

Theorem \ref{thm:recurrence} illustrates the recurrence of a compact set $\mathcal{R}$, together with the condition $\partial \mathcal{R} \cap \Omega(f) = \emptyset$, necessarily implies its containment within the region of attraction of $\Omega(f)\cap\R$. As a result, by imposing mild conditions on $\Omega(f)$, one leads to the following quite useful result.

{\color{black}
\begin{corollary}\label{cor:roa-subet}
Let assumptions \ref{as:roa} hold.
Further, let $\R$ be a compact set satisfying $\partial\R\cap\Omega(f)=\emptyset$ and $\Omega(f)\cap\R=\{x^*\}$. Then the set $\R$ is recurrent if and only if $\R\subset\A(x^*)$.
\end{corollary}
\ifthenelse{\boolean{with-appendix}}{
}
{
\begin{proof}
($\implies$):~
By assumption $\R$ is compact, and $\partial\R\cap\Omega(f)=\emptyset$. Then, Theorem \ref{thm:recurrence} implies that if $\R$ is recurrent then $\Omega(f)\cap\R\not=\emptyset$ and $\R\subset\A(\Omega(f)\cap\R)$. This, together with the assumption that $\Omega(f)\cap\R=\{x^*\}$, implies $\R\subset\A(x^*)$.

($\Longleftarrow$):~
This direction is trivial given Theorem~\ref{thm:recurrence}.
\end{proof}
}
}
Corollary \ref{cor:roa-subet} implies that from a practical standpoint, one may use recurrence as a mechanism for finding inner approximations for $\A(x^*)$. However, one limitation of the above results is that although $\R$ is recurrent, we do not know a priori how long it may take for a trajectory to come back to $\R$ after it leaves it. This motivates the following stricter notion of recurrence.

\begin{definition}[$\tau$-Recurrent Set]\label{defn:T recurrent}
A  set $\mathcal{R} \subseteq \mathbb{R}^d$ is $\tau$-recurrent w.r.t. \eqref{eq:system}, \textcolor{black}{if for any point $x_0\in \R$ and any time $t\ge 0$, there exists a $t'\in (t,t+\tau]$, such that $\phi(t',x_0)\in \R$.}
\end{definition}

\begin{theorem}\label{thm:bounded k}
Let Assumption \ref{as:roa} hold, and consider a compact set $\R\subseteq \mathcal{A}(x^*)$ satisfying $x^*\in \opint \R$ and $\R\cap \partial \A(x^*)=\emptyset$.
Then there exists positive constants $\underline{c}$\,, $\overline{c}$\,, and $a$\,, depending on $\R$, such that for all  
$\tau\geq \bar \tau:=\frac{\overline{c}-\underline{c}}{a}$,
the set $\R$ is $\tau$-recurrent. Further, starting from any point $x\in \R$, the solution $\phi(t,x)\in \R$ for all $t\ge \bar \tau$.
\end{theorem}

\begin{figure}
    \center
    \includegraphics[width=0.25\textwidth]{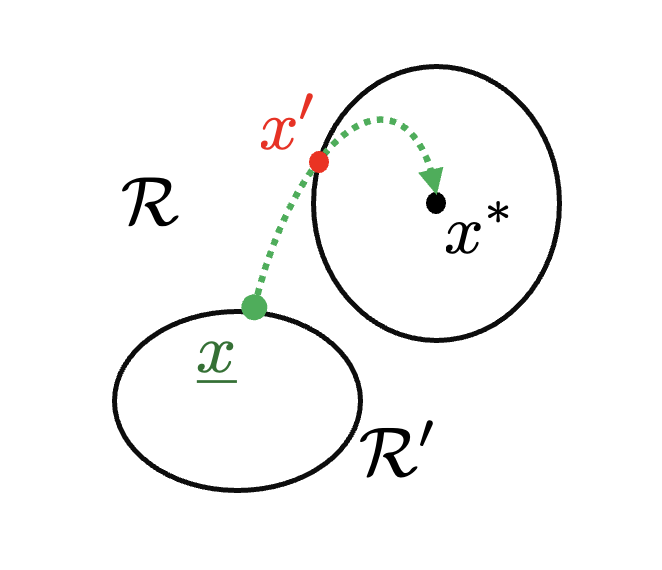}
    
    \caption{{\color{black}An visualization of the proof of Theorem~\ref{thm:bounded k}. Particularly, when $\mathcal{R}$ is disconnected, the equilibrium $x^*$ must be contained in $\mathcal{R}'$, since otherwise, one could find a point $x'\in\partial\mathcal{R}$ with $V(x')<V(\underline{x})$ along the trajectory $\phi(t,\underline{x})$ that strictly decreases $V$. }}
    \label{fig:disconnected}
\end{figure}

\ifthenelse{\boolean{with-appendix}}{
}{
\begin{proof}
The proof of the theorem relies on Zubov's existence criterion stated in Theorem \ref{thm:zubov}. Given $\R$, let us now define
\begin{align}
    \underline{c} := \min_{x\in\partial\R} V(x), \quad \overline{c} := \max_{x\in\partial\R} V(x), \\
    \text{and}\quad a := \max_{x\in C} \nabla V(x)^T f(x),
\end{align}
where $C=\{x\in\mathbb{R}^d: \underline{c}\leq V(x)\leq\overline{c}\}$ is compact. 

We first argue that $V_{\leq\underline{c}}:=\{x:V(x)\le \underline{c}\}\subseteq \R$. Let $\underline{x}$ be the point in $\partial\R$ that achieves the minimum, i.e, $V(\underline{x})=\underline{c}$. Since $\R$ is not necessarily connected, we use $\R'$ to denote the connected component of $\R$ containing $\underline{x}$. Note that $x^*\in\opint \R$ must be contained in $\R'$, since otherwise, the trajectory $\phi(t,\underline{x})$, which strictly decreases $V$ must eventually find a point $x'\in\partial R$ with $V(x')<\underline{c}$; which contradicts the  definition of $\underline{c}$, see Fig~\ref{fig:disconnected}. Thus, $x^*\in\R'\subseteq\R$. 

Suppose then that $V_{\leq \underline{c}}\not\subseteq\R'\subseteq\R$, for any point $\tilde x\in V_{\leq \underline{c}}\backslash\R'$, $V(\phi(t,\tilde x))<\underline{c}$, for $t>0$, and $\lim_{t\rightarrow\infty}\phi(t,\tilde x)=x^*$. Thus there exists $\tilde t>0$ s.t. $V(\phi(\tilde t,\tilde x))<\underline{c}$ and $\phi(\tilde t,\tilde x)\in\partial \R$; which contradicts again with the  definition of $\underline{c}$. It follow then that $V_{\leq\underline{c}}\subseteq \R' \subseteq \R$.

Similarly, since the contradictable set $V_{\leq\overline{c}}$ contains every point in the boundary of $\R$, there cannot be any point in $x\in\R$ with $V(x)>\overline{c}$. 
We therefore get that the following inclusions must hold:
\begin{equation}\label{eq:inclusion}
V_{\leq\underline{c}}\subseteq \R \subseteq V_{\leq\overline{c}}.
\end{equation}

Finally, by \eqref{eq:inclusion}, for any point $x\in \R$ we must have $V(x)\leq \overline{c}$. Since the
time derivative of $V(x)$ is at most $a<0$, it follows that \textcolor{black}{after $t\ge \Bar{\tau}:=\frac{\underline{c}-\overline{c}}{a}$ the Lyapunov value $V(\phi(t,x))\leq \underline{c}$, which implies that $\phi(t,x)\in\R$ and result follows.}
\end{proof}
}

Note that the lower bound on $\tau$ in Theorem~\ref{thm:bounded k} implicitly depends on the set ${\mathcal{R}}$. This makes the process of learning a recurrent set difficult as $\tau$ would change, and the set is updated.
To eliminate this dependence, one is required to introduce conservativeness. To that end, for given $\delta>0$, $c\in(0,1)$, and $V$ as in Theorem \ref{thm:zubov}, we consider the set
\textcolor{black}{
\begin{align}\label{eq:r_delta}
        \tilde{\A}:={V_{\le {c}}}\backslash\{\opint\mathcal{B}_\delta+x^*\},
\end{align}
where as mentioned before $V_{\leq{{c}}}:=\{x:V(x)\le {{c}}\}$ is a compact Lyapunov sublevel set contained in $\A(x^*)$ .} The sign '$+$' in \eqref{eq:r_delta} represents the Minkowski sum, and $\mathcal{B}_\delta$ is a closed $\delta$ ball centered at the origin, i.e., $\mathcal{B}_\delta = \{x|\norm{x}_2\le\delta\}$. Note we further choose $\delta>0$ to be small enough such that \textcolor{black}{$\mathcal{B}_{\delta}+x^*\subseteq{V_{\le {c}}}$, and the set $V_{\le {c}}$ can approximate the ROA $\A(x^*)$ with arbitrary ($2$-norm) accuracy as ${c}\rightarrow1$ in the case that $\A(x^*)$ is bounded.} 

\textcolor{black}{
Then, by denoting $\underline{c}(\delta)$ as the min Lyapunov function value in $\tilde{\mathcal{A}}$, and $a(\delta)$ as the largest Lie derivative within the set $C_\delta=\{x\in\mathbb{R}^d: \underline{c}(\delta)\leq V(x)\leq {c}\}$, i.e.,
\begin{align}\label{eq:delta-params}
    \underline{c}(\delta) := \min_{x\in\tilde{\mathcal{A}}} V(x),\quad\text{and}\quad 
    a(\delta) := \max_{x\in C_\delta} \nabla V(x)^T f(x),
\end{align}
we obtain a lower bound on $\tau$ that is independent of $\R$.}

\textcolor{black}{\begin{theorem}\label{thm:recurrent subset}
Let Assumption~\ref{as:roa} hold, and consider $\delta>0$, $c\in(0,1)$ and a compact set  ${\mathcal{R}}$ satisfying:
$\mathcal{B}_\delta+x^* \subseteq{\mathcal{R}}\subseteq V_{\le {c}}$. Then $\R$ is $\tau$-recurrent for $\tau\ge \bar{\tau}(\delta):=(\underline{c}(\delta)-{c})/a(\delta)$. Moreover, when $t\ge \bar{\tau}(\delta)$, $\phi(t,x)\in \R$ for any point $x\in \R$.
\end{theorem}
}
\ifthenelse{\boolean{with-appendix}}{}{
\begin{proof}
    \textcolor{black}{
    Let us first construct a contradiction to show $V_{\le \underline{c}(\delta)}\subseteq \mathcal{B}_\delta+x^*$. Particularly, if $V_{\le \underline{c}(\delta)}\not\subseteq \mathcal{B}_\delta+x^*$, then for any point $\tilde{x}\in V_{\le \underline{c}(\delta)}\backslash\{\mathcal{B}_\delta+x^*\}$, $\lim_{t\rightarrow\infty}\phi(t,\tilde{x})=x^*$ and $V(\phi(t,\tilde{x}))<\underline{c}(\delta)$ for all $t>0$. Therefore, there exists a $\tilde t>0$ such that $V(\phi(\tilde t,\tilde{x}))<\underline{c}(\delta)$ and $\phi(\tilde t,\tilde{x})\in\partial\{\mathcal{B}_\delta+x^*\}\subset \tilde{\mathcal{A}}$, which contradicts with the definition of $\underline{c}(\delta)$.}
    
    \textcolor{black}{
    Now, since $V_{\le \underline{c}(\delta)}\subseteq\mathcal{B}_\delta+x^* \subseteq{\mathcal{R}}\subseteq V_{\le {c}}$, any point $x\in\mathcal{R}$ must have $V(x)\le c$. Then, it follows from the definition of $a(\delta)$ that after $t\ge\overline{\tau}(\delta)$, the Lyapunov value $V(\phi(t,x))\le\underline{c}(\delta)$, and thus $\phi(t,x)\in\mathcal{R}$.
    }
\end{proof}
}



\section{Learning recurrent sets}\label{sec:Solution Approach}
Having laid down the basic theory underlying recurrent sets, we now propose a method to compute inner approximations of the region of attraction $\A(x^*)$ based on checking the recurrence property on finite-length trajectory samples. 
For concreteness, we consider the following type of sampled trajectories for system \eqref{eq:system}:
\begin{align}\label{eq:dt-system}
    x_n = \phi(n \tau_s,x_0), \qquad x_0\in\mathbb{R}^d\,, \quad n\in\mathbb{N}\, ,
\end{align}
where $\tau_s>0$ is the sampling period.  

In this setting, we define the notion of discrete-time recurrence w.r.t. a length $k$ trajectory:
\begin{definition}[$k$-Recurrent Set]\label{defn:k-step recurrent}
A set $\mathcal{R} \subseteq \mathbb{R}^d$ is $k$-steps recurrent (k-recurrent for short) w.r.t. \eqref{eq:dt-system}, \textcolor{black}{if for any point $x_0\in \R$ and any step index $n\ge0$, there exists an $n'\in \{n+1,...,n+k\}$, such that $x_{n'}\in \R$.}
\end{definition}

\begin{remark}
Note that a set $\R$ being $k$-recurrent implies that $\R$ is $\tau$-recurrent with $\tau=k\tau_s$. One can then conclude that $\R\subset \mathcal{A}(x^*)$ under the assumptions of Corollary~\ref{cor:roa-subet}. However, the converse is not necessarily true.
\end{remark}
To ensure one can find such a $k$-recurrent set, we consider again the specific set $\tilde{\A}$ defined in \eqref{eq:r_delta} that gives the following sufficient conditions for a set $\R$ to be $k$-recurrent. 

\begin{theorem}\label{thm:k-recurrent subset}
\textcolor{black}{
Let Assumption~\ref{as:roa} hold, and consider $\delta>0$, $c\in(0,1)$ and a compact set  ${\mathcal{R}}$ satisfying:
$\mathcal{B}_\delta+x^* \subseteq{\mathcal{R}}\subseteq V_{\le {c}}$. Then $\R$ is $\tau$-recurrent for  $k>\Bar{k}(\delta):=\Bar{\tau}(\delta)/\tau_s$, where $\Bar{\tau}(\delta)$ is defined as in Theorem~\ref{thm:recurrent subset}.}
\end{theorem}
\ifthenelse{\boolean{with-appendix}}{}{
\begin{proof}
Given Theorem~\ref{thm:recurrent subset}, this result  follows directly from $\phi(t,x)\in \R$ for all $x\in\R$ when $t\ge\Bar{\tau}(\delta)$.
\end{proof}
}

 In the rest of the paper, we assume w.l.o.g. that the asymptotically stable equilibrium is at the origin, i.e., $x^* =0$. We briefly explain next the underlying mechanism that will be used to learn recurrent sets.

\noindent
 \paragraph*{Algorithm Summary}
 We will restrict our search to a compact initial  approximation $ \hat{\mathcal{S}}^{(0)}\subset \mathbb{R}^d$ of the ROA satisfying $\hat{\mathcal{S}}^{(0)}\supseteq\mathcal{B}_\delta$. Precisely, we will seek to find a subset of the ROA within $\A(x^*)\cap\hat{\mathcal{S}}^{(0)}$ by computing $k$-recurrent sets $\R$ that seek to satisfy the properties of Theorem \ref{thm:k-recurrent subset}.
In this approach, starting from $ \hat{\mathcal{S}}^{(0)}$, we sequentially generate a sequence of approximations $\hat{\mathcal{S}}^{(i)}$. For each $\hat{\mathcal{S}}^{(i)}$, we sample points $p_{ij}\in \hat{\mathcal{S}}^{(i)}$ and check whether a trajectory of length $k$ that starts at $p_{ij}$ returns to $\hat{\mathcal{S}}^{(i)}$ for each $j=0,1,2,...$. A trajectory that does not return to $\hat{\mathcal{S}}^{(i)}$ within $k$ steps is a counter-example of $k$-recurrence.
Once a counter-example is found, we update the approximation $\hat{\mathcal{S}}^{(i)}$ to $\hat{\mathcal{S}}^{(i+1)}$ and restart the sampling process. This method is illustrated in Algorithm \ref{alg:1}.

The rest of this section provides a detailed explanation of each step of the algorithm, as well as a rigorous justification of the proposed methodology.

\begin{algorithm}
\caption{Learning a $k$-recurrent set}\label{alg:1}

    Initialize $\hat{\mathcal{S}}_0$ according to \eqref{eqn:c_shpere}\\
    \For{Iteration $i=0,1,...$}{
    \For{Iteration $j=0,1,...$}{
    Generate random sample $p_{ij}\in\hat{\mathcal{S}}^{(i)} $ uniformly\\
  \If{$p_{ij}$ is a counter-example w.r.t $\hat{\mathcal{S}}^{(i)}$}{
    {Update $\hat{\mathcal{S}}^{(i)}$ according to \eqref{eq:update sphere}}\\
    {\textbf{break}}
    
   }
   }
  }
\end{algorithm}

\subsection{Classification of sample points}

We say that a sample point $p_{ij}$ is a valid $k$-recurrent point w.r.t current approximation $\hat{\mathcal{S}}^{(i)}$ if starting from $x_0=p_{ij}$, 
\begin{align}
\exists\,\,n\in \{1,...,k\}, \,\, {s.t.} \,\, x_{n}\in \hat{\mathcal{S}}^{(i)}.
\label{eq:positive-ex-criterion}
\end{align}
If \eqref{eq:positive-ex-criterion} does not hold, we say $p_{ij}$ is a \emph{counter-example}. We will use such counter-examples to update our current set approximation $\hat{\mathcal{S}}^{(i)}$.

\subsection{Construction of set approximations}
In order to gradually update the sets $\hat{\mathcal{S}}^{(i)}$, we consider two parametric families of set approximations.
\subsubsection{Sphere approximation}
To construct a sphere approximation, we start by 
choosing a radius $\bar b>0$ large enough such that the set 
\begin{align}\label{eqn:c_shpere}
\hat{\mathcal{S}}^{(0)}:=\{x|\norm{x}_2\le b^{(0)}:={\color{black} \bar b}\} \;\supseteq\; \mathcal{B}_{\delta}.
\end{align}
The sphere approximation for iteration $i$ is then defined as $\hat{\mathcal{S}}^{(i)} := \{x|\norm{x}_2\le b^{(i)}\}$.
Finally, given a sample point $p_{ij}\in \hat{\mathcal{S}}^{(i)}$ we update $\hat{\mathcal{S}}^{(i)}$ based on the following criterion:
\begin{align}\label{eq:update sphere}
    \textbf{$p_{ij}$ counter-example }& \!\!\implies b^{(i+1)} = \norm{p_{ij}}_2-\varepsilon,\quad
\end{align}
where $\varepsilon>0$ is an algorithm parameter expressing the level of conservativeness in our update. 

If the process reaches a value of $b^{(i)}<\delta$, we declare the search a failure. At such point, one may choose to either reduce the value of $\varepsilon$ or increase the length of the trajectories sampled.

\subsubsection{Polyhedron approximation}
One could also choose to construct $\hat{\mathcal{S}}^{(i)}$ using polyhedron approximation. To that end, we first construct a matrix $A:=[a_1,...,a_n]^T\in \mathbb{R}^{n\times d}$, where each row vector  $a_n$ is a \emph{normalized} ($\|a_l\|=1$ $\forall l$) exploration direction indexed by $l\in\{1,...,n\}$. Precisely, we generate $A$ randomly with row dimension $n$ large enough such that for any arbitrary direction $v\in\mathbb{R}^d$, there exists an exploration direction $a_{l}$ with an angle $\theta_{v,l}$ between $v$ and $a_{l}$ satisfying
\begin{align}\label{eq:r net}
    \theta_{v,l}:=\arccos{\left(\frac{v^Ta_l}{\|v\|\|a_l\|}\right)}\le \frac{2}{3}\pi.
\end{align}
Note that the aforementioned process is analog to constructing an $\varepsilon$-net over a unit Euclidean hyper-sphere, for which several algorithms exist~\cite{Haussler1987netsAS,mustafa2019}. Upper bounds on the size of $n$ can also be found in the literature, see, e.g., \cite{vershynin2011introduction}.

The polyhedron approximation at iteration $i$ is defined as $\hat{\mathcal{S}}^{(i)} := \{x|Ax\le b^{(i)}\}$, consisting on $n$ inequalities aimed at approximating a $k$-recurrent set via counter examples. In this paper, keep $A$ fixed for every iteration, and update the constraint coefficients $b^{(i)} = [b^{(i)}_1,...,b^{(i)}_n]^T\in\mathbb{R}^n$.

Similarly to the sphere approximation, we initialize $b^{(0)}:=\textcolor{black}{\bar b} \mathbbm{1}_n$, such that
\begin{align}\label{eqn: c_poly}
\hat{\mathcal{S}}^{(0)}=\{x|Ax\le b^{(0)}:=\textcolor{black}{\bar b} \mathbbm{1}_n\} \supseteq \mathcal{B}_{\delta},
\end{align}
which can be done for $\textcolor{black}{\bar b}>0$ large enough; here $\mathbbm{1}_n\in\mathbb{R}^n$ is the vector of all ones. 

Afterwards $b^{(i)}$ is updated by sampling points $p_{ij}\in \hat{\mathcal{S}}^{(i)}$ and checking the following criterion:
\begin{align}\label{eq:update polyhedron}
    &\textbf{$p_{ij}$ counter-example}\nonumber\\
    &\implies\,\,\begin{cases}
        b^{(i+1)}_{l^*} =a_{l^*} p_{ij}-\varepsilon\\
        b^{(i+1)}_l = b^{(i)}_l, \quad \forall\,l = \{1,..,n\}\backslash l^*,
    \end{cases}
\end{align}
where $\varepsilon>0$ is fixed and $l^* = \argmax_{l\in \{1,...,n\}}  \frac{a_l^T p_{ij}}{\|a_l\|\|p_{ij}\|}$ is the index of exploration direction that minimizes the angle between $p_{ij}$ and $a_l$. If $l^*$ consists of more than one index, we simply choose one at random. 
As before, we declare the search a failure whenever $b_l^{(i)}<0$ for some $l$, since this implies that the equilibrium $x^*=0$ is outside the set $\hat{\mathcal{S}}^{(i)}$.

\ifthenelse{\boolean{with-appendix}}{}{
\begin{figure}
    \center
    \includegraphics[width=0.33\textwidth]{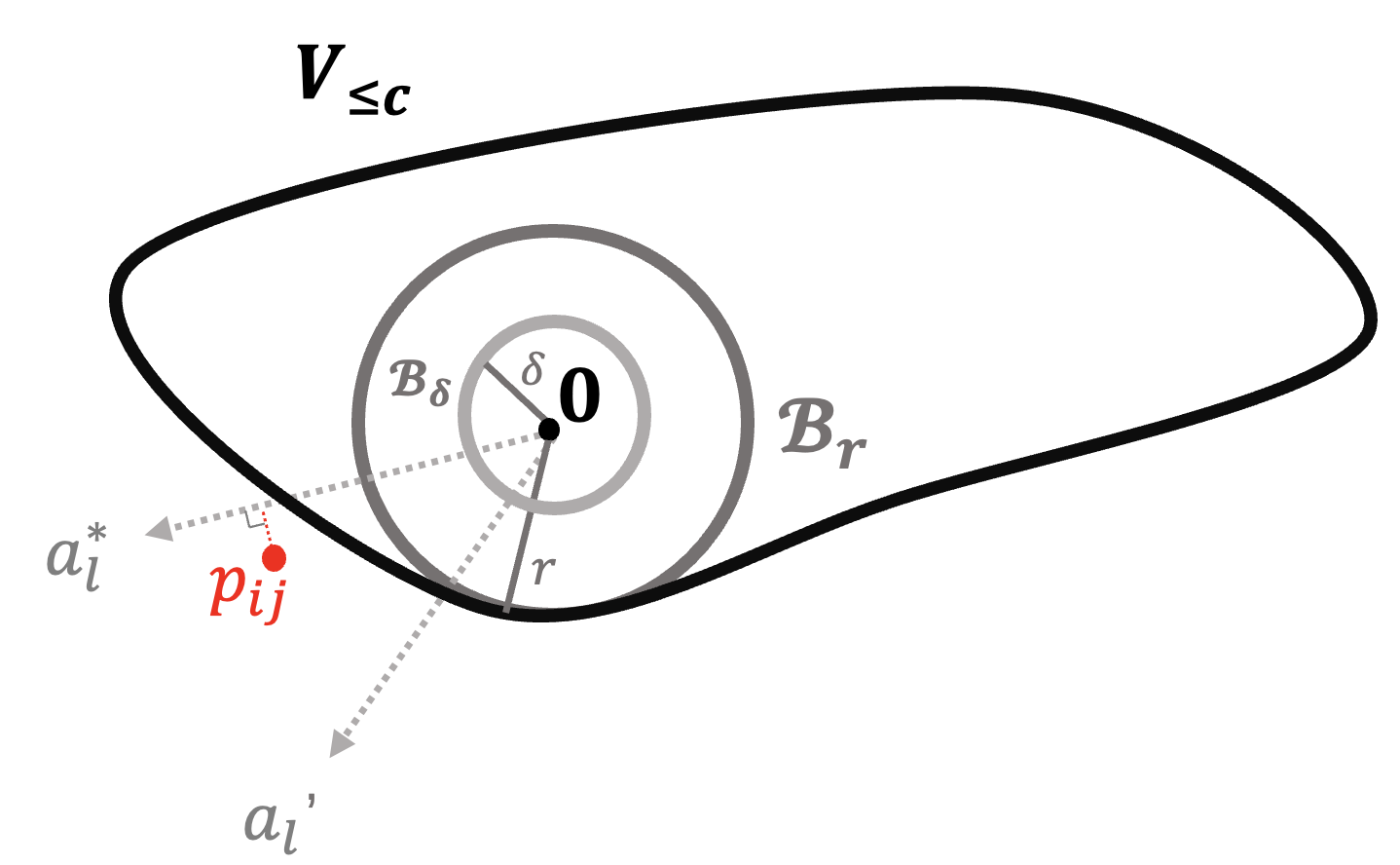}
    \caption{An illustration of the proof of Theorem~\ref{thm: inclusion}. In particular, given an arbitrary point $p_{ij} \not \in  \textcolor{black}{ V_{\le {c}}}$, in the sphere case, it follows that $\norm{p_{ij}}_2\ge r$. And in the polyhedron case, the closest projection  $\max_{l\in \{1,...,n\}} {a_l^T p_{ij}} \ge r/2$ under Assumption~\eqref{eq:r net} that every pair of exploration directions are close enough.}
    \label{fig:inclu thm}
\end{figure}
}

\subsection{Bound on the number of updates}
As mentioned before, the aforementioned search for approximations will fail if $b^{(i)}$ (sphere) or one of $b^{(i)}_l$ (polyhedron) becomes negative \textcolor{black}{at some iteration $i>0$}. We will show next that, provided that $k$ and $\varepsilon$ are chosen appropriately, there will be no failure. In other words, there will be no counter-examples after a finite number of set updates.

Let us recall $\Bar{k}(\delta)$ defined in Theorem~\ref{thm:k-recurrent subset}. Then, given $k\ge \Bar{k}(\delta)$, and an arbitrary approximation $\Hat{\mathcal{S}}^{(i)}$ satisfying \textcolor{black}{$\mathcal{B}_\delta \subseteq\Hat{\mathcal{S}}^{(i)}\subseteq V_{\le c}$}, Theorem~\ref{thm:k-recurrent subset} guarantees that any sample $p_{ij}\in \Hat{\mathcal{S}}^{(i)}$ will lead to a $k$-recurrent trajectory, i.e., condition \eqref{eq:positive-ex-criterion}. As a result, the algorithm will stop updating at this point since we cannot find further counter-examples within $\hat{\mathcal S}^{(i)}$.
This means that, if it is possible for $\Hat{\mathcal{S}}^{(i)}$ to become a subset of \textcolor{black}{$V_{\le c}$}, without violating the condition $\mathcal{B}_\delta \subseteq\Hat{\mathcal{S}}^{(i)}$, then the algorithm will stop updating and will never fail. The following theorem shows that this is indeed the case, whenever $\varepsilon$ and $k$ are properly chosen.

\begin{theorem}\label{thm: inclusion}
Let the initial approximation $\hat{\mathcal{S}}^{(0)}$ satisfy $\mathcal{B}_\delta\subseteq\hat{\mathcal{S}}^{(0)}$ and trajectory length $k>\Bar{k}(\delta)$, for $\bar k(\delta)$ as defined in Theorem \ref{thm:k-recurrent subset}. Then, given a counter-example $p_{ij}$, the resulting updated set satisfies $\mathcal{B}_\delta\subseteq\hat{\mathcal{S}}^{(i+1)}$ whenever
\begin{align}\label{eq:epsilon}
    \varepsilon \le \begin{cases}
    r-\delta \quad &\textbf{with sphere approximation}\\
     \frac{r}{2}-\delta\quad &\textbf{with polyhedron approximation},
    \end{cases}
\end{align}
where $r$ is the smallest distance between the origin (equilibrium) and the boundary \textcolor{black}{$\partial V_{\le {c}}$}.
\end{theorem}
\ifthenelse{\boolean{with-appendix}}{
}{
\begin{proof}
Given an arbitrary counter-example $p_{ij}$ w.r.t $\hat{\mathcal{S}}^{(i)}\supseteq\mathcal{B}_\delta$, it follows that $p_{ij} \not \in \textcolor{black}{ V_{\le {c}}}$ by Theorem~\ref{thm:k-recurrent subset}; since otherwise, $p_{ij}$ would generate a $k$-recurrent trajectory. Then, \textcolor{black}{it follows from the definition of $r$ that  }$\norm{p_{ij}}_2\ge r$, as illustrated in Figure \ref{fig:inclu thm}. Further, let $\mathcal{B}_r:= \{x|\norm{x}_2 \le r\}\subseteq  \textcolor{black}{ V_{\le {c}}}$. 

We now reason differently depending on the type of approximation.
\vspace{-3ex}
\begin{description}[align=left,style=nextline,leftmargin=*,labelsep=\parindent,font=\normalfont]
\item \emph{(Sphere case):}~
It then follows from $\norm{p_{ij}}_2\ge r$ that whenever $\varepsilon\le r-\delta$, the update leads to $\,b^{(i+1)} = \norm{p_{ij}}_2-\varepsilon \ge r-\varepsilon \ge \delta$.\vspace{-2.5ex}
 \item \emph{(Polyhedron case):}~
 It follows from~\eqref{eq:r net}, that for any point $p' \not\in \mathcal{B}_r$, we have $\max_{l\in \{1,...,n\}}  a_l^T p'\geq \|p'\|\cos \left(\frac{2}{3}\pi\right)\geq \frac{r}{2}.$
 Therefore, since by definition of $\mathcal B_r$, $p_{ij}\not\in \mathcal B_r$ we conclude then that
 $b^{(i+1)}_{l^*} =a_{l^*}^T p_{ij}-\varepsilon\ge \frac{r}{2} - \varepsilon\geq \delta.$
\end{description}
Together with the fact that $\hat{\mathcal{S}}^{(0)}\supseteq \mathcal{B}_\delta$, result follows.
\end{proof}
}


Theorem \ref{thm: inclusion} establishes that one can choose parameters $k$ and $\varepsilon$ so that the sequence of sets $\hat{\mathcal{S}}^{(i)}$ never leads to $b^{(i)}$ or $b_l^{(i)}$ negative, i.e., the algorithm never fails. However, this requires prior knowledge of $\bar k(\delta)$, $r$, and $\delta$.
We argue that local information on the dynamics can be sufficient to find conservative bounds for $r$ and $\delta$, and thus $\varepsilon$. 
However, $\bar k(\delta)$ depends in a highly non-trivial way on $\delta$. We solve this issue by, doubling the side of $k$, i.e. $k^+=2k$, every time the failure conditions are met, and re-initializing the sets back to $\hat{\mathcal{S}}^{0}$. 

In what follows, we use {\color{black}$\mathcal F_{\bar b}$} to denote the parametric family of closed balls (resp. polytopes) defined by $\{x:||x||_2\leq b\}$ (resp.$\{x:Ax\leq b\}$), for $b\in[0,{\bar b}]$ (resp. $b\in [0,{\bar b}]^n$). This leads to the following total bound on the number of iterations.

 
\begin{theorem}\label{thm:counter example upper bound}
Given the initial approximation $\hat{\mathcal{S}}^{(0)}\in \mathcal{F}_{\bar{b}}$ and initial constant $\bar{b}$ defined in \eqref{eqn:c_shpere} or \eqref{eqn: c_poly}, the total number of counter-examples encountered in Algorithm~\ref{alg:1}, with $k$-doubling after each failure, is bounded by $\frac{\bar b}{\varepsilon}\log_2{\Bar{k}(\delta)}$ in the sphere case and $n\frac{\bar b}{\varepsilon}\log_2{\Bar{k}(\delta)}$ in the polyhedron case.
\end{theorem}
\ifthenelse{\boolean{with-appendix}}{}{
\begin{proof}
 Note that once a counter-example is encountered, we decrease the radius constraint (sphere case) or one of the exploration directions (polyhedron case) by at least $\varepsilon$. Therefore, $\hat{\mathcal{S}}^{(i)}\in \mathcal{F}_{\bar{b}}$ for all $i\in\{1,2,...\}$. And for any fixed $k$, our method can find at most $\bar{b}/\varepsilon$ counter-examples with the sphere approximation and $n\bar{b}/\varepsilon$ counter-examples with the polyhedron approximation without failing. Since it takes at most $\log_2{\Bar{k}(\delta)}$ updates on $k$ to find some $k\geq \bar k(\delta)$ using the doubling method, result follows.
\end{proof}
}

Our results provide an upper bound on the number of updates the set approximation may experience by ensuring that $\hat{\mathcal{S}}^{(i)}$ always contains an $\delta$-ball around the equilibrium point. However, this is not sufficient to guarantee that $\hat{\mathcal{S}}^{(i)}$ is $k$-recurrent, which is required to guarantee that \textcolor{black}{$\hat{\mathcal{S}}^{(i)}\subseteq V_{\le c}$}. This issue is addressed next.

\subsection{Convergence guarantee}
By Definition~\ref{defn:k-step recurrent}, a set $\hat{\mathcal{S}}$ is $k$-recurrent if every point $p \in \hat{\mathcal{S}}$ satisfies \eqref{eq:positive-ex-criterion}. As shown before, certifying this property will enable us to guarantee that $\hat{\mathcal{S}}\subset\A(0)$. However, it is infeasible to enforce condition \eqref{eq:positive-ex-criterion} for every point in $\hat{\mathcal{S}}$. 
Instead, we will show that under mild conditions, our algorithm converges to a $\hat{\mathcal{S}}^*$ satisfying $\opint\hat{\mathcal{S}}^*\subseteq \A(0)$ with probability one.

In our algorithm, we generate samples $p_j$ uniformly within some set $\hat{\mathcal{S}}$, i.e., $p_j\stackrel{iid}{\sim} U(\hat{\mathcal{S}})$ for all $j\in\{0,1,2,...\}$. We use $\hat{\mathcal{S}}_{\text{counter}}$ to denote the set that contains all the counter-examples  \textcolor{black}{ in $\hat{\mathcal{S}}$} that certify  $\hat{\mathcal{S}}$ being not $k$-recurrent\textcolor{black}{, i.e., 
\begin{align}
    \hat{\mathcal{S}}_{\text{counter}}: = \{p\in \hat{\mathcal{S}}\vert \,p \text{ is a counter-example}\}.
\end{align}}Given a random sample $p_j$ we define the Bernoulli random variable $X_j$ with $X_j = 1$ if $p_j\in \hat{\mathcal{S}}_{\text{counter}}$ and $X_j = 0$ otherwise.

{\color{black}
\begin{lemma}\label{lem:int not recurrent}
Consider a set $\hat{\mathcal{S}}\in \mathcal F_{\bar b}$ satisfying $\partial \mathcal{\hat{S}}\cap \Omega(f) = \emptyset$ and $\Omega(f)\cap\hat{\mathcal{S}}=\{0\}$, if $\opint\hat{\mathcal{S}}\not\subseteq\mathcal{A}(0)$, then there exists a point $p\in \opint\hat{\mathcal{S}}\backslash\mathcal{A}(0)$ and a time $t'>0$ such that $\phi(t,p)\not\in \hat{\mathcal{S}}$, $\forall t>t'$. Moreover, the point $p$ could be selected such that $t'$ is arbitrarily close to zero.
\end{lemma}
\ifthenelse{\boolean{with-appendix}}{}{
\begin{proof}
Let us consider a point $p\in \opint\hat{\mathcal{S}}\backslash\mathcal{A}(0)$, we claim that either $p\not\in \mathcal{A}(\Omega(f))$ or $p\in \mathcal{A}(\Omega(f)\backslash \{0\})$ is true since $ \mathcal{A}(\Omega(f)\backslash \{0\})\cap \mathcal{A}(0) = \emptyset$.

Consider first a point $p\not\in \mathcal{A}(\Omega(f))$, we have $\phi(p,t) \rightarrow \infty$ as $t \rightarrow \infty$. Since $\hat{\mathcal{S}}$ is compact, there must exists a time $t'>0$ such that $\phi(t,p)\not\in \hat{\mathcal{S}}$, $\forall t>t'$ in this case. 

In the other case that $p\in \mathcal{A}(\Omega(f)\backslash \{0\})$, it follows from the definition of the regions of attraction (Definition~\ref{defn:ROA}) that $\liminf_{t\rightarrow\infty} d(\phi(t,p), \Omega(f)\backslash \{0\})=0$. Note that by assumption $\hat{\mathcal{S}}$ is a compact set and $\{\Omega(f)\backslash \{0\}\}\cap \hat{\mathcal{S}} = \emptyset$. Therefore, if the result does not follow, i.e., for all $t'>0$ there exists a $t> t'$ such that $\phi(t,p)\in\hat{\mathcal{S}}$, we can construct an infinite sequence  $\{x_{n}\}_{n=0}^\infty\subset\partial\hat{\mathcal{S}}$ as in the proof of Lemma~\ref{lem:bounded traj}. Then, since $\partial\hat{\mathcal{S}}$ is compact, by Bolzano-Weierstrass theorem, $\{x_{n}\}_{n=0}^\infty$ must have a sub-sequence  $\{x_{n_i}\}_{i=1}^\infty$ that converges to an accumulation point $\bar x\in \partial\hat{\mathcal{S}}$. It follows then from the definition of $\omega$-limit sets (Definition \ref{defn:omega-limit-set}) that $\bar x= \lim_{i\rightarrow \infty}x_{n_i}\in \Omega(f)\cap\partial\hat{\mathcal{S}}$, which contradicts with the assumption that $\partial\hat{\mathcal{S}}\cap\Omega(f)=\emptyset$.

Now, since the first result follows, we can additionally let $q=\phi(t'-\delta,p)$ and conclude  $\phi(t,q)\not\in \hat{\mathcal{S}}$, $\forall t> \delta$ with $\delta>0$ arbitrarily close to zero, thus $t$ is arbitrarily close to zero. 
\end{proof}
}

\begin{lemma}\label{lem:scountervol}
For any set  $\hat{\mathcal{S}}\in \mathcal F_{\bar b}$ satisfying $\partial \mathcal{\hat{S}}\cap \Omega(f) = \emptyset$ and {\color{black}$\Omega(f)\cap\hat{\mathcal{S}}=\{0\}$}, the volume of its counter-example part is positive, i.e., $\vol(\hat{\mathcal{S}}_\text{\emph{counter}})>0$, whenever $\opint\hat{\mathcal{S}}\not\subseteq\mathcal{A}(0)$.
\end{lemma}
\ifthenelse{\boolean{with-appendix}}{}{
\begin{proof}
If $\opint\hat{\mathcal{S}}\not\subseteq\mathcal{A}(0)$, then Lemma~\ref{lem:int not recurrent} implies that there exists a point $p\in \opint\hat{\mathcal{S}}\backslash\mathcal{A}(0)$ and a time $t' \in (0, \tau_s)$ such that $\phi(t,p)\not\in \hat{\mathcal{S}}$, $\forall t> t'$. Then, for any $k$, we can respectively construct a neighborhood of $p$ that consists of counter-examples of $k$-recurrent.

Precisely, let us consider an arbitrary point $q \in \mathcal{\hat{S}}$ and recall the assumption that the dynamical system \eqref{eq:system} is globally $L$-Lipschitz, it follows from  \cite[p. 96]{Khalil2002} that the distance between solutions $\norm{\phi(t,p)-\phi(t,q)}\le\norm{p-q}\exp{(Lt)}$ for all $t\ge0$. Therefore, we choose $q$ such that 
\[
\norm{p-q}<\underline{d}:=\min_{n=1,...,k} d(\phi(n \tau_s,p),\mathcal{\hat{S}})/\exp{(Lk \tau_s)},
\]
and claim $\norm{\phi(n\tau_s,p)-\phi(n\tau_s,q)}\le\norm{p-q}\exp{(Ln\tau_s)}<d(\phi(n\tau_s,p),\mathcal{\hat{S}}),\,\,\forall n=1,...,k$, i.e., $q$ is a counter-example of $k$-recurrence.

Finally, since $p\in\opint \mathcal{\hat{S}}$, the aforementioned counter-example set $\{q \in \mathcal{\hat{S}}| \norm{p-q}<\underline{d}\}$
has positive volume, and thus the result follows.
\end{proof}
}
}


\begin{lemma}\label{lem:counter example as}
For any set $\hat{\mathcal{S}}\in\mathcal{F}_{\bar b}$ satisfying $\opint \hat{\mathcal{S}}\not\subseteq\mathcal{A}(0)$, {\color{black}$\partial \mathcal{\hat{S}}\cap \Omega(f) = \emptyset$ and $\Omega(f)\cap\hat{\mathcal{S}}=\{0\}$}, we have $\lim_{m\rightarrow\infty}  \mathbb{P}(X_1=\dots=X_m=1)=0.$
That is, a counter-example is eventually sampled almost surely.
\end{lemma}
\ifthenelse{\boolean{with-appendix}}{}{
\begin{proof}
Note that we have $\hat{\mathcal{S}}_{\text{counter}}\subseteq\hat{\mathcal{S}}$ and $\vol(\hat{\mathcal{S}}_\text{counter})>0$ by Lemma~\ref{lem:scountervol}. Then, denoting the counter-example ratio as $\rho:=\vol(\hat{\mathcal{S}}_\text{counter})/\vol(\hat{\mathcal{S}})$, one can conclude $0<\rho\le1$ and 
\begin{align}
    \lim_{m\rightarrow\infty}\mathbb{P}(X_0=...=X_m=1)=\lim_{m\rightarrow\infty}(1-\rho)^{m} = 0.
\end{align}
\end{proof}
}

We now leverage the results in Lemma \ref{lem:scountervol} and Lemma \ref{lem:counter example as} to obtain the following termination guarantee.

\begin{theorem}\label{thm: Sas}
Consider $\hat{\mathcal{S}}^{(0)}\in\mathcal{F}_{\bar b}$ with  $\hat{\mathcal{S}}^{(0)}\supseteq \mathcal{B}_{\delta}$ and {\color{black}$\Omega(f)\cap\hat{\mathcal{S}}^{(0)}=\{0\}$}. Then, after at most $\frac{\bar b}{\varepsilon}\log_2{\Bar{k}(\delta)}$ (resp. $n\frac{\bar b}{\varepsilon}\log_2{\Bar{k}(\delta)}$) iterations in the sphere (resp. polyhedron) case, the updates on  $\hat{\mathcal{S}}^{(i)}$ terminate at some $\hat{\mathcal{S}}^*$ whose interior is a non-empty subset of $\A(0)$ whenever $k>\Bar{k}(\delta)$ and \eqref{eq:epsilon} holds.
\end{theorem}
\ifthenelse{\boolean{with-appendix}}{}{
\begin{proof}
  Suppose that at any given iteration $i$ the set $\opint \hat{\mathcal{S}}^{(i)}\not\subseteq\mathcal{A}(0)$. Then it follows from Lemma \ref{lem:counter example as} that a counter-example is eventually found almost surely, and a new set $\hat{\mathcal{S}}^{(i+1)}$ is obtained.
  Also Theorem~\ref{thm:counter example upper bound} implies the total number of such transitions is upper bounded by $\frac{\bar b}{\varepsilon}\log_2{\Bar{k}(\delta)}$ (resp. $n\frac{\bar b}{\varepsilon}\log_2{\Bar{k}(\delta)}$) in the sphere (resp. polyhedron) case, since $\hat{\mathcal{S}}^{(0)}\in \mathcal{F}_{\bar b}$ and $\mathcal{B}_\delta \subseteq\hat{\mathcal{S}}^{(0)}$.
  
  Now let $\hat{\mathcal{S}}^*$ denote the last updated approximation. Note that since there are not further updates to  $\hat{\mathcal{S}}^*$  with probability one, this implies that $\vol(\hat{\mathcal{S}}^*_{\text{counter}})=0$. We argue then that $\opint \hat{\mathcal{S}}^{*}\subseteq\mathcal{A}(0)$, since otherwise Lemma~\ref{lem:scountervol} implies $\vol(\hat{\mathcal{S}}^*_{\text{counter}})>0$, which contradicts the fact that $\hat{\mathcal{S}}^{*}$ is the last iteration. Finally, $\opint \hat{\mathcal{S}}^*$ is non-empty since Theorem~\ref{thm: inclusion} implies $\hat{\mathcal{S}}^*\supseteq\mathcal{B}_\delta$.

\end{proof}
}
\begin{figure*}[h]
\centerfloat
\includegraphics[width=1.05\linewidth]{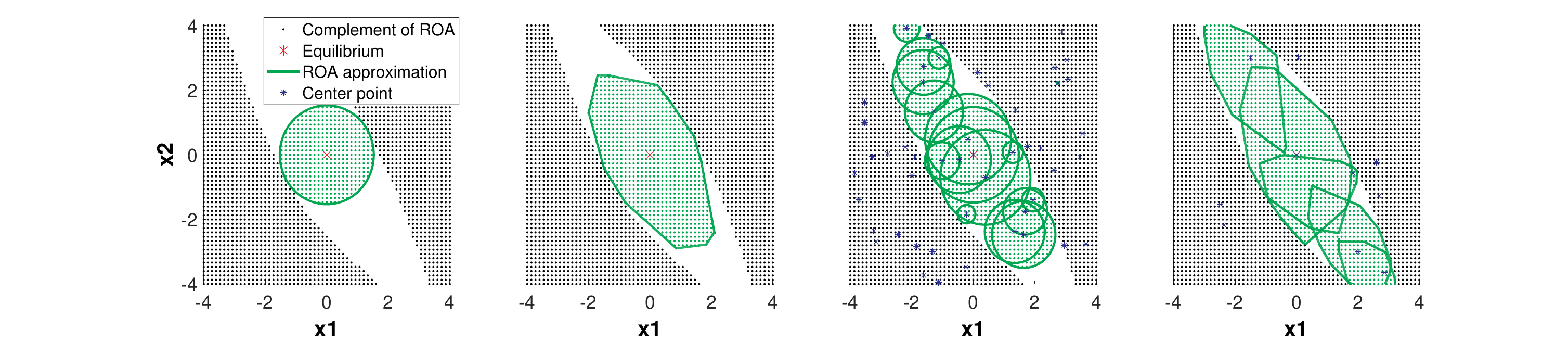}
\caption{The region of attraction approximations for one center point (left two) and multiple random center points (right two) in the sphere and polyhedron case, respectively.}
\label{fig: example}
\end{figure*}

\subsection{Multiple center point approximation}
When the ROA $\mathcal{A}(0)$ is distorted or non-convex, Algorithm~\ref{alg:1} may significantly underestimate the set $\mathcal{A}(0)$, meaning that the volume of the resulting approximation $\vol (\mathcal{\hat{S}}^{(i)})\ll\vol (\mathcal{A}(0))$. 
To address this problem, we can refine Algorithm~\ref{alg:1} by generating additional approximations similar to $\mathcal{\hat{S}}^{(i)}$ but centered at points different from the equilibrium $x^*=0$. 

In particular, we consider $h\in\mathbb{N}^+$ center points $x_q$ indexed by $q \in \{ 1,2,...,h\}$, where the first center point as $x_1=x^*=0$. Then other centers, i.e., $x_2$,...,$x_h$, can be chosen uniformly within some region of interest or selected to be in some preferred place. At each center point $x_q$ the sphere approximation is defined by $\mathcal{\hat{S}}^{(i)}_q:=\{x\vert \norm{x-x_q}_2 \le b^{(i)}_q\}$, where $b^{(i)}_q$ represents the radius to be updated in the presence of counter-examples. As before we initialize $b^{(0)}_q=\bar b$. In the case of polyhedral approximations, we similarly define $b_q^{(i)} = [b^{(i)}_{q,1},...,b^{(i)}_{q,n}]^T\in\mathbb{R}^n$, with $b^{(0)}_q= \bar{b} \mathbbm{1}^N$ and let $\hat{S}^{(i)}_q:=\{x\vert A(x-x_q)\le b^{(i)}_q\}$.

Then, the multi-center ROA approximation $\hat{\mathcal{S}}_{\text{multi}}^{(i)}$ at iteration $i$ is the union of all approximations, i.e., $\hat{\mathcal{S}}_{\text{multi}}^{(i)} := \cup_{q=1}^{h} \mathcal{\hat{S}}^{(i)}_q$. Note that $\mathcal{\hat{S}}^{(i)}_1$ is equivalent to the original approximation $\mathcal{\hat{S}}^{(i)}$ of previous sections, and $\mathcal{\hat{S}}^{(i)}_2$ to $\mathcal{\hat{S}}^{(i)}_h$ are additional enhancements.

 Similar to Algorithm~\ref{alg:1}, sample points $p_{ij}$ are generated uniformly within $\hat{\mathcal{S}}_{\text{multi}}^{(i)}$ in each sub-iteration $j=1,2,...$. In this multi-center case, $p_{ij}$ is classified as a counter-example if starting from $x_0 = p_{ij}$, $x_n \not\in \hat{\mathcal{S}}_{\text{multi}}^{(i)}$ for all $n\in\{1,...,k\}$. Once encountered a counter-example, we update $\hat{\mathcal{S}}_{\text{multi}}^{(i)}$ and restart sampling iteration $j$.  
 In particular, given a counter-example $p_{ij}\in \hat{\mathcal{S}}_{\text{multi}}^{(i)}$, every approximations $\hat{\mathcal{S}}^{(i)}_q$ (sphere or polyhedron) satisfying $p_{ij} \in \hat{\mathcal{S}}^{(i)}_q$ are subjected to update respectively via the following criterion:
 \begin{subequations}\label{eq:update addition}
 \begin{align}
     \textbf{(sphere) }\,&\quad b_q^{(i+1)}= \norm{p_{ij}-x_q}_2-\varepsilon 
     \\
    \textbf{(polyhedron) }&\begin{cases}
        b^{(i+1)}_{q,l^*} =a_{l^*} (p_{ij}-x_q)-\varepsilon\\
        b^{(i+1)}_{q,l} = b^{(i)}_{q,l},\, \forall\,l\! \in\! \!\{1,..,n\}\!\backslash l^*,
    \end{cases}
 \end{align}
 \end{subequations}
  where $l^* = \argmax_{l\in \{1,...,n\}}  \frac{a_l^T (p_{ij}-x_q)}{\norm{a_l}\norm{p_{ij}-x_q}} $. Again, we choose one at random if $l^*$ consists of more than one index.

  Then, those approximations not containing $p_{ij}$ are updated as $\hat{\mathcal{S}}^{(i+1)}_q=\hat{\mathcal{S}}^{(i)}_q$. Note that the parameter $\varepsilon$ is strictly positive. Thus, for all center points $x_q\not\in\mathcal{A}(0)$,  the corresponding constraint parameters $b_q^{(i)}$ could decrease to negative values and result in   $\hat{\mathcal{S}}^{(i)}_q=\emptyset$ without affecting our results.

In this multi-center setting, we use $\mathcal F^h_{\bar b}$ to denote the parametric family of $h$ closed balls (resp. polytopes) defined by $\cup_{q=1}^h \mathcal{S}_q$, where $\mathcal{S}_q = \{x:||x-x_q||_2\leq b_q\}$ (resp. $\mathcal{S}_q = \{x:A(x-x_q)\leq b_q\}$), for $b_q\in[0,{\bar b}]$ (resp. $b_q\in [0,{\bar b}]^n$) and $x_q\in\mathbb{R}^d$ indexed by $q=\{1,...,h\}$.
  
  \begin{theorem}\label{thm: multi center}
  For any iteration $i\in\mathbb{N}^+$, the multi-center approximation $\hat{\mathcal{S}}_{\text{\emph{multi}}}^{(i)}$ is non-vanishing, i.e., $\hat{\mathcal{S}}_{\text{\emph{multi}}}^{(i)}\supseteq\mathcal{B}_\delta$, if $k>\bar k$ and condition \eqref{eq:epsilon} is satisfied. The total number of counter-examples encountered, with $k$-doubling after each failure, is bounded by $h\frac{{\bar b}}{\varepsilon}\log_2{\Bar{k}(\delta)}$ and $nh\frac{{\bar b}}{\varepsilon}\log_2{\Bar{k}(\delta)}$ in the sphere and polyhedron case respectively. Moreover, the last updated multi-center approximation $\hat{\mathcal{S}}_{\text{\emph{multi}}}^{*}$ satisfies $\opint \hat{\mathcal{S}}_{\text{\emph{multi}}}^{*}\not=\emptyset$ and $\opint \hat{\mathcal{S}}_{\text{\emph{multi}}}^{*}\subseteq\mathcal{A}(0)$  {\color{black} whenever $\Omega(f)\cap\hat{\mathcal{S}}^{(0)}_{\text{multi}}=\{0\}$}.
  \end{theorem}
  \ifthenelse{\boolean{with-appendix}}{}{
  \begin{proof}
  By definition $\hat{\mathcal{S}}_{\text{multi}}^{(i)}\supseteq\hat{\mathcal{S}}_1^{(i)}$ for all $i\in\mathbb{N}^+$, Theorem~\ref{thm: inclusion} therefore implies $\hat{\mathcal{S}}_{\text{multi}}^{(i)}\supseteq\hat{\mathcal{S}}_1^{(i)}\supseteq\mathcal{B}_\delta$ under \eqref{eq:epsilon}. The bound on the total number of counter-examples follows as in Theorem~\ref{thm:counter example upper bound}, since every additional approximation $\hat{\mathcal{S}}_q^{(i)}\in \mathcal{F}_{\bar b}$ for all $q\in\{1,...,h\}$ and iteration $i\in\{1,...\}$. Finally, by generalizing Lemma ~\ref{lem:int not recurrent}-~\ref{lem:counter example as} and Theorem~\ref{thm: Sas} to the scope of $\mathcal{F}_{\bar b}^h$, the last statement follows.
  \end{proof}}


\begin{table*}[htp]
\centerfloat
 \begin{tabular}{||c| c| c| c| c||} 
 \hline
  Approximate method& \# of counter examples& \# of samples & \# of steps simulated& Average \# of steps per sample  \\ [0.5ex] 
 \hline\hline
 1-center sphere approximation & 14 & 7024 & 7935&1.39 \\ \hline
 1-center polyhedron approximation & 94 & 23130 & 28127&1.22 \\
 \hline
 50-center sphere approximations & 191 & 17481 & 53756&3.07 \\
 \hline
 10-center polyhedron approximations & 370 & 46819 & 66399&1.41 \\ [1ex] 
 \hline
 \end{tabular}
 \caption{Performance statistics for different configurations of our algorithm.}
 \label{table:1}
 \end{table*}
 
\section{Experiments}\label{sec:experiments}

We illustrate the accuracy of the proposed  methodology by approximating the region of attraction of the following autonomous dynamical system:
\begin{align}\label{eq:experiment dynamics}
    \begin{bmatrix}
    \dot{x}_1\\\dot{x}_2
    \end{bmatrix}=
    \begin{bmatrix}
    x_2\\-x_1+\frac{1}{3}x_1^3-x_2
    \end{bmatrix}.
\end{align}
The black dotted area in Figure~\ref{fig: example} represents the complement of ROA of the origin, which is computed by testing a mesh grid of points. A point is marked black if it does not converge to the equilibrium after $t=30$. In our algorithm we set $\epsilon=0.1$, $k=50$ and $\tau_s=0.5$. To estimate the number of iterations until convergence, we stop our algorithm when all black dots are excluded from our current approximation.

The outcomes of our approximation are marked in green. In particular, Figure \ref{fig: example} (left two panels) shows the outcome of applying Algorithm~\ref{alg:1} using a sphere and a $n=200$ directions polyhedron approximation. To address the problem of under-estimation, as shown in the right two panels of Figure~\ref{fig: example}, we can generate random center points. Fifty spheres or ten polyhedrons approximates with random center points give a good approximation. Detailed statistics of our algorithms for the aforementioned scenarios are provided in Table~\ref{table:1}. 
Notably, the number of counter-examples and the steps simulated per sample is small, which illustrates the efficiency of our algorithm.

\section{Conclusions and future work}\label{sec:conclusions}
We consider the problem of learning the region of attraction of a stable equilibrium point. We propose the use of a more flexible notion of invariance known as recurrence. We provide necessary and sufficient conditions for a recurrent set to be an inner approximation of the ROA. Our algorithms are sequential and only incur a limited number of counter-examples. Future work includes extending our framework to other families of approximations and control design.

\balance

\bibliographystyle{IEEEtran}
\bibliography{refs.bib}
\ifthenelse{\boolean{with-appendix}}{
\clearpage
\input{sections/appendix}
}{}
\end{document}